\documentclass[letterpaper, 10 pt, conference]{ieeeconf}  

\IEEEoverridecommandlockouts                              

\usepackage{lipsum} 
\newcommand{\N}{\mathbb{N}}
\newcommand{\R}{\mathbb{R}}

\newcommand{\act}{\mathbf{a}}
\newcommand{\obs}{\mathbf{o}}
\newcommand{\obsrob}{\obs^{\mathrm{robot}}} %
 
\newcommand{\obsdloact}{\obs^{\mathrm{dlo},\pi}} %
\newcommand{\obsdlocri}{\obs^{\mathrm{dlo,V}}} %
\newcommand{\obstas}{\obs^{\mathrm{task}}} %

\newcommand{\go}{\mathbf{p}}

\newcommand{\pos}{\mathbf{p}}
\newcommand{\vel}{\dot{\mathbf{p}}}

\newcommand{\NDR}{\emph{N+DR}}
\newcommand{\NDRD}{\emph{N+DR+D}}
\usepackage[dvipsnames]{xcolor}
\usepackage{etoolbox}
\usepackage{letltxmacro}
\usepackage{booktabs} 
\usepackage{array} 
\usepackage{capt-of} 
\usepackage{placeins} 
\usepackage[percent]{overpic}
\usepackage[font=footnotesize,labelfont=bf]{caption}
\usepackage{adjustbox}
\usepackage{subcaption}
\usepackage{multirow}
\usepackage{float}
\newcommand{\snapshotlabel}[1]{%
    \colorbox{white}{%
        \textcolor{black}{\scriptsize\bfseries #1}%
    }%
}
\usepackage{xcolor}
\usepackage{pifont}
\usepackage{tabularx}
\let\labelindent\relax
\usepackage[inline]{enumitem}
\definecolor{CiteHL}{RGB}{255,245,204} 
\usepackage[colorlinks=true, urlcolor=blue, linkcolor=black, citecolor=BrickRed]{hyperref}
\usepackage[normalem]{ulem} 
\makeatletter
\newcommand{\CiteHL@box}[1]{\begingroup\setlength\fboxsep{1pt}\colorbox{CiteHL}{#1}\endgroup}
\AtBeginDocument{%
  \LetLtxMacro{\CiteHL@origcite}{\cite}
  \newcommand{\CiteHL@cite@opt}[2][]{\CiteHL@box{\CiteHL@origcite[#1]{#2}}}%
  \newcommand{\CiteHL@cite@noopt}[1]{\CiteHL@box{\CiteHL@origcite{#1}}}%
  \renewcommand{\cite}{\@ifnextchar[{
    \CiteHL@cite@opt
  }{
    \CiteHL@cite@noopt
  }}%
}
\makeatother

\usepackage{graphics} 
\usepackage{epsfig} 
\usepackage{mathptmx} 
\usepackage{times} 
\usepackage{amsmath} 
\usepackage{amssymb}  
\usepackage{cuted}
\title{\LARGE \bf
An Empirical Study of Reinforcement Learning for Keypoint Positioning of Deformable Linear Objects} 

\author{Vinay Rodrigues$^{1,2}$, Sara Zeynalpour$^1$, Gokhan Solak$^1$, Sebastian Hjorth$^1$, Arash Ajoudani$^1$%
\thanks{$^{1}$Human-Robot Interfaces and Interaction (HRII) Laboratory, Istituto Italiano di Tecnologia (IIT), 16163 Genoa, Italy.}%
\thanks{$^{2}$Ph.D. Program of National Interest in Robotics and Intelligent Machines (DRIM), and Universit\`a degli Studi di Genova, Genoa, Italy.}%
\thanks{{\tt\small vinayvivian.rodrigues@iit.it}}%
\thanks{This work was supported by the European Union Horizon Project TORNADO (GA 101189557).}%
}

\begin{document}

\maketitle
\thispagestyle{empty}
\pagestyle{empty}

\begin{abstract}
Deformable linear object (DLO) manipulation is challenging because high-dimensional configurations, nonlinear contact dynamics, and diverse material and geometric properties produce a broad range of object responses that are both difficult to model and data-intensive to learn. This work studies reinforcement learning for planar manipulation of a one-end-grasped DLO, focusing on training efficiency, observation design, robustness, and zero-shot sim-to-real transfer. 
We compare three Proximal Policy Optimization training stacks and find that the GPU-parallel MuJoCo Playground stack enables faster convergence and more consistent performance than CPU-based alternatives under the same wall-clock budget. 
Relative keypoint observations and an asymmetric actor–critic formulation further improve learning efficiency. Policies trained with observation noise, delay, domain randomization, and disturbances achieve simulation success rates of 96.4\% for single-point tracking, 98.0\% for disturbance recovery, and 88.8\% for two-point tracking. Using markerless RGB-D keypoint estimation, the policies transfer without real-world fine-tuning, succeeding in 9/10, 10/10, and 8/10 physical trials, respectively.
\end{abstract}
\section{Introduction}
\label{sec:introduction}
Deformable linear objects (DLOs), such as cables, hoses, and ropes, are common in industrial and
domestic environments. The ability to manipulate such objects is important for enabling humanoid robots to assist with tasks, such as cable routing~\cite{qin2023dual} and installation~\cite{qin2019cable}. In contrast to rigid objects, a DLO cannot be represented by a single pose. Motion applied at one grasp point may induce complex deformations throughout the object.

Additionally, reliable DLO manipulation is challenging due to the object's high-dimensional
configuration, nonlinear dynamics, frictional contact, self-collision, and
sensitivity to geometric and material properties. In real-world
settings, these difficulties are compounded by visual occlusion, measurement
noise, perception latency, uncertain physical parameters, and external
disturbances. For robotic systems, the object-level control
problem must integrate with robot kinematics, and potentially whole-body motion~\cite{qin2023dual}. 

To address these challenges, DLO manipulation has been investigated through
physical modeling, geometric control, planning, and learning-based
methods~\cite{yin2021modeling}. Model-free reinforcement learning (RL) is particularly attractive in
this setting because it can acquire closed-loop manipulation behavior without
requiring an explicitly identified DLO dynamics model.
\begin{figure}[t]
    \centering
    \begin{overpic}[width=0.9\columnwidth,trim={5.5cm 5.5cm 7.8cm 4.1cm},clip]{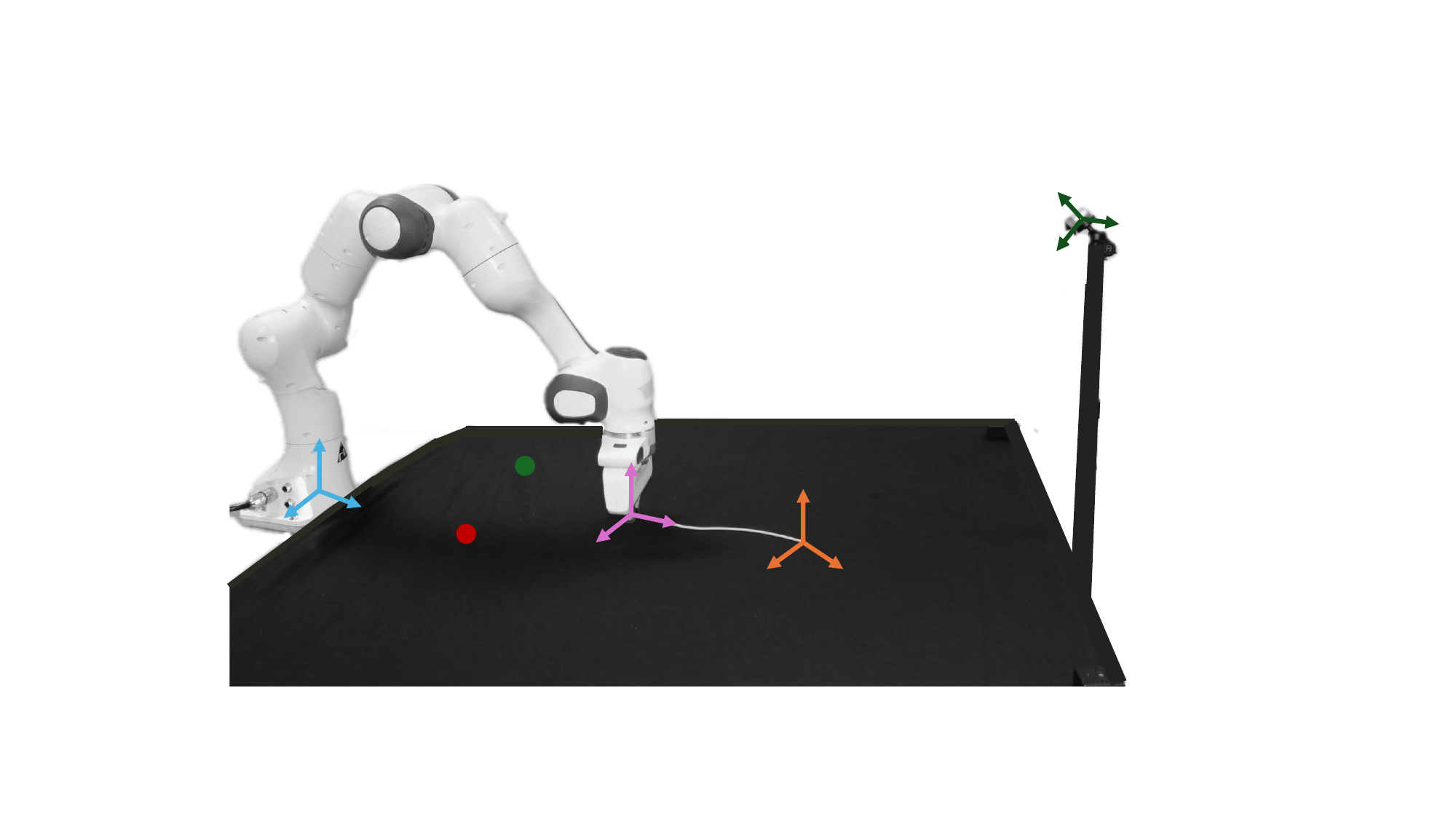}
        \put(65,5){\textcolor{orange}{$\pos^0_{k_{\mathrm{tip}},t}.$}}
        \put(47,8.5){\textcolor{Lavender}{$\pos^0_{k_{\mathrm{att}},t}$}}
        \put(27.5,10){\textcolor{ForestGreen}{$\go^0_{G,k_{\mathrm{att}}}$}}
        \put(26.2,3.5){\textcolor{BrickRed}{$\go^0_{G,k_{\mathrm{tip}}}$}}
        \put(2.6,11){\textcolor{Cyan}{$\mathbf{\{0\}}$}}
        \put(87,40){\textcolor{Green}{$\mathbf{\{C\}}$}}
    \end{overpic}
    \caption{Planar one-end-grasped DLO setup and target geometry.}
    \label{fig:task_description}
    \vspace{-5mm}
\end{figure}
%
Laezza et al.\cite{laezza2021shape} used DDPG to learn simulated
shape control of elastoplastic DLOs, while Zakaria et al.\cite{zakaria2022robotic} formulated DLO manipulation as
moving selected object nodes toward desired three-dimensional positions using
Cartesian robot actions. Daniel et al.\cite{daniel2024multiac} decomposed translational and rotational end-effector (EE) control between multiple actor--critic agents and demonstrated large three-dimensional deformations and transfer across physical DLOs. To reduce the reliance
on online interaction, Laezza et al.\cite{laezza2024offline} investigated
offline goal-conditioned RL for reaching previously unseen target shapes from
a limited real-world dataset. Although these studies demonstrate the effectiveness of RL for DLO
control, they do not jointly investigate the practical efficiency of the complete training system, single- and
multi-waypoint tracking, one-policy operation across multiple DLO lengths,
disturbance recovery, and vision-based closed-loop deployment. This work
addresses these aspects within a unified model-free RL framework.

Training efficiency is particularly important for model-free RL because policy optimization may require millions of environment interactions and repeated runs across hyperparameters, random seeds, and task variations. Prior comparisons of RL implementations have shown that JAX-based learners can improve throughput over conventional PyTorch implementations, while also indicating that the achievable gain is limited when physics simulation remains the dominant computational bottleneck~\cite{nico_rlx}. This motivates evaluating not only the learning backend, but also frameworks that jointly accelerate policy optimization and parallel physics simulation on GPUs, such as MuJoCo Playground (MP)~\cite{zakka2025mujocoplayground}.

In practice, however, training systems differ in several coupled components, including the learning implementation, the simulation backend, the hardware execution, and the number of parallel environments. Their performance therefore cannot be interpreted as a controlled comparison of individual learning algorithms or simulation methods alone. Instead, this work evaluates their practical performance as complete end-to-end training stacks for DLO manipulation.

The task considered in this work is Cartesian control of a DLO whose one end is attached to the robot EE. The DLO is represented by task-relevant keypoints, and the policy maps robot proprioception, DLO state, and target-relative information to Cartesian EE increments. This task-space formulation is independent of a particular robot joint representation and can be interfaced with robot-specific inverse-kinematics, operational-space, or whole-body controllers.

Three complete PPO training stacks are tuned, trained, and evaluated: 
PyTorch Stable-Baselines3 (SB3) and Stable-Baselines Jax (SBX)\cite{stable-baselines3}, both coupled with CPU-based MuJoCo\cite{todorov2012mujoco}, and the integrated JAX implementation in MP \cite{zakka2025mujocoplayground}. Since the three training stacks differ not only in the learning implementation
but also in the simulation backend and hardware execution, they are compared
in terms of both wall-clock efficiency and held-out task performance. Based on
this comparison, MP is selected for the remaining experiments. We then evaluate
relative and robot-base-frame DLO observations, symmetric and asymmetric
actor--critic formulations, single- and two-point tracking, robustness to
observation noise and delay, external disturbances, and training across three
DLO lengths. Finally, the learned task-specific policies are transferred without real-world fine-tuning and executed using markerless RGB-D keypoint estimation. The main contributions can be summarized as:
\begin{itemize}
    \setlength{\itemsep}{0pt}
    \setlength{\parskip}{0pt}
    \setlength{\parsep}{0pt}
    \item a systematic characterization of the computational bottlenecks in model-free DLO manipulation, separating the effect of an accelerated learning backend from that of massively parallel physics simulation through a held-out comparison of SB3+MuJoCo, SBX+MuJoCo, and MP;
    \item a controlled ablation of the RL formulation, quantifying the effects of DLO representations and actor--critic observations on learning efficiency, followed by an evaluation of robustness to observation noise and delay, disturbances, and variations in DLO geometry; and
    \item zero-shot transfer of the learned task-specific policies to a physical robot using a markerless \hbox{RGB-D} perception system,
    with evaluation on single-point tracking and disturbance recovery, and two-point tracking.
\end{itemize}
%
%
%
%
%
%
%
%
%
\section{RL for DLO Manipulation}
\label{sec:method}
This section presents the closed-loop framework used to learn and deploy
policies for planar DLO manipulation. The manipulation setup and closed-loop architecture are introduced first, followed by the RL framework and task setup, vision pipeline, task formulation, and training strategy.

\subsection{Manipulation Setup and Closed-Loop Architecture}
\label{subsec:system_overview}

\begin{figure}[t]
    \centering
    \includegraphics[width=0.75\columnwidth,trim={19.25cm 1cm 0.8cm 1.8cm},clip]{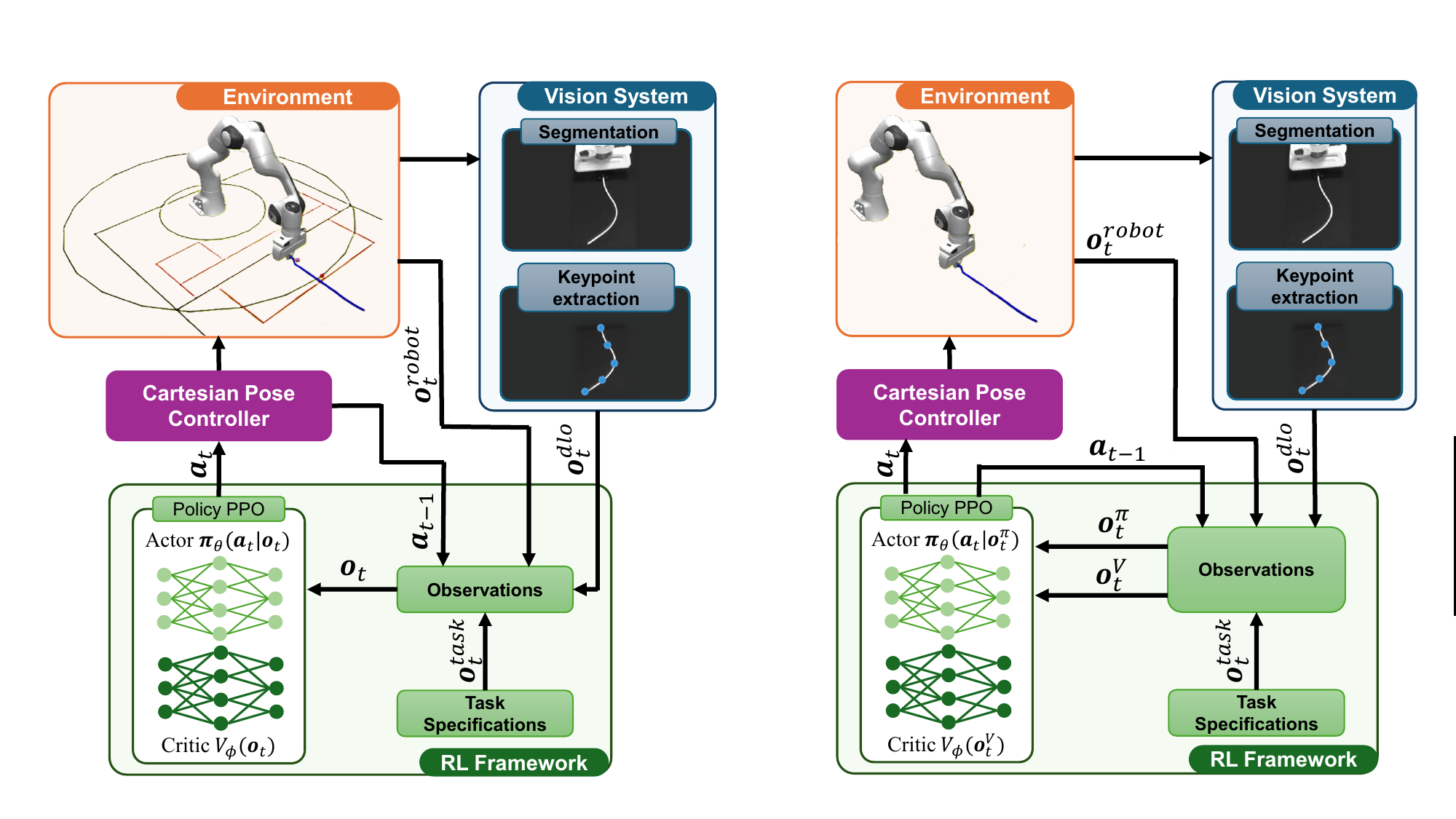}
    \caption{Closed-loop deployment and asymmetric actor–critic training.}
    \label{fig:framework}
    \vspace{-5mm}
\end{figure}

The closed-loop perception--action architecture is shown in
Fig.~\ref{fig:framework}. At control step $t$, robot proprioception is obtained
directly from the robot or simulator. The actor-observable DLO keypoints are
read from the simulator during training and estimated from RGB-D input during
real-world deployment. These quantities are combined with the task
specification $\obstas_t$ and the previous action $\act_{t-1}$ to construct the
actor observation $\mathbf{o}^{\pi}_t$. The action $\act_t$ is converted by the Cartesian pose controller into an
EE motion command. The resulting robot and DLO states are used to construct the next
observation, thereby closing the loop. During asymmetric actor--critic training, the
critic additionally receives privileged information from the simulator, whereas the
deployed controller requires only the actor-observable inputs. The observation
and action spaces are defined in Sec.~\ref{subsub:tdef}.
%
\subsection{Vision System}\label{subsec:vision}
\begin{figure}
    \centering
    \includegraphics[width=.95\columnwidth,trim={0.7cm 4.8cm 12.75cm 5.75cm},clip]{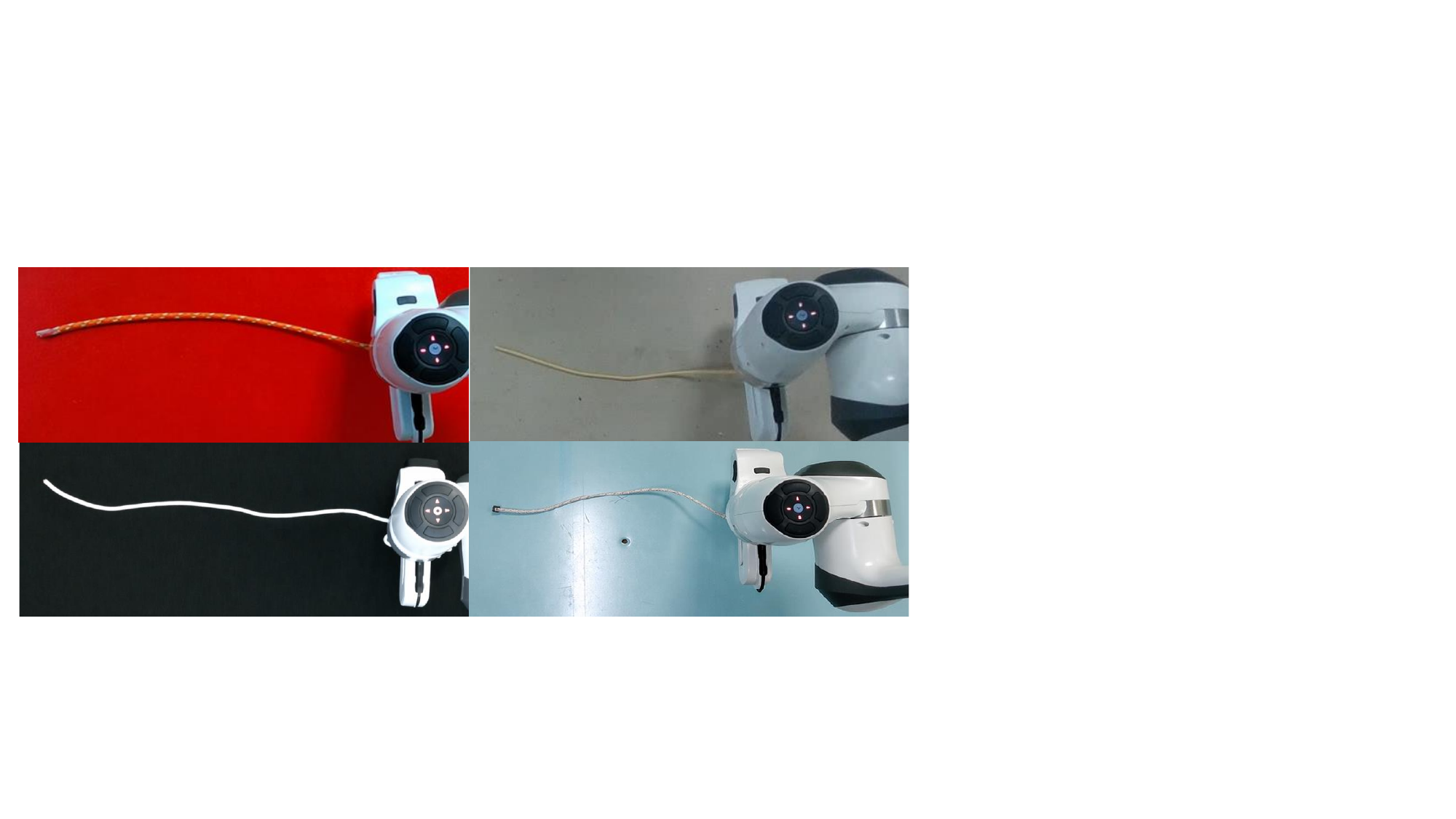}
    \caption{Examples from the task-specific rope perception dataset used to fine-tune the \textit{YOLO26-seg} model. 
}
    \label{fig:vision_insieme}
    \vspace{-5mm}
\end{figure}

The proposed policy requires the online state of the DLO, which is estimated using a markerless RGB-D vision pipeline. 
Since the original system was developed for a different manipulation scenario, we collected a task-specific dataset to improve robustness across different rope types, backgrounds, lighting conditions, viewpoints, and the presence of distractor objects such as the robot itself (Fig.~\ref{fig:vision_insieme}). 

A YOLO26-seg model was fine-tuned~\cite{jocher2026ultralytics} to segment the rope in each RGB image. The predicted mask is skeletonized, and $N\in\N$ approximately equally spaced keypoints are extracted along the rope centerline to obtain a compact representation of the DLO state following ~\cite{long2026self}.  Depth projection gives camera-frame points $\pos^{\mathrm{c}}_{i} \in \mathbb{R}^{3}$
, ordered from the grasped end to the free end of the rope; then they are projected onto the table plane in 
robot-base frame $\pos^{\mathrm{0}}_{i} \in \mathbb{R}^{2}$.

This markerless perception pipeline provides a practical alternative to motion-capture systems and fiducial markers while enabling real-time state estimation for closed-loop manipulation. Additional implementation details and quantitative evaluation are provided in Sec.~\ref{subsec:real_world_protocol}.
%
\subsection{Task Definition and RL Framework}
\label{subsub:tdef}
All planar quantities are expressed in the robot-base frame. The DLO manipulation problem is formulated as an episodic Markov decision
process (MDP)
$\mathcal{M}=(\mathcal{S},\mathcal{A},p,r,\gamma,\rho_0)$.
The actor receives a compact
observation compatible with the information available from the robot and
vision system, whereas the critic receives additional privileged simulator
information during training. Let $\mathcal{I}_V=\{1,\ldots,N_V\}$ denote the ordered index set of DLO keypoints available in simulation, arranged from the EE-attached endpoint, $k_{\mathrm{att}}=1$, to the free tip, $k_{\mathrm{tip}}=N_V$. The actor observes only an ordered subset
$\mathcal{I}_{\pi}\subseteq\mathcal{I}_V$, with
$|\mathcal{I}_{\pi}|=N_{\pi}$ and $\{k_{\mathrm{att}},k_{\mathrm{tip}}\}\subseteq\mathcal{I}_{\pi}$. The actor receives the observation
\begin{equation}
    \mathbf{o}^{\pi}_t =
    \left[
        \obsrob_t,\,
        \obsdloact_t,\,
        \obstas_t,\,
        \act_{t-1}
    \right]
    \in\R^{13+4N_{\pi}+2M},
    \label{eq:actor_observation}
\end{equation}
where $M$ is the number of task-relevant keypoints and
$\act_{t-1}\in\R^3$ is the previous action. The robot proprioceptive observation is
$\obsrob_t=
[\pos^{0}_{\mathrm{EE},t},\,
\vel^{0}_{\mathrm{EE},t},\,
\cos\psi_t,\,
\sin\psi_t,\,
\dot{\psi}_t,\,
\eta_t]\!\in\R^{8}$, where $\pos^{0}_{\mathrm{EE},t},\vel^{0}_{\mathrm{EE},t}\in\R^2$
denote the planar EE position and velocity, $\psi_t,\dot{\psi}_t\in\R$ denote the EE yaw and yaw rate, and
$\eta_t\in[0,1]$ is the normalized radial distance of the EE from the robot base. For keypoint $i$, the position and velocity relative to the EE are defined as $\pos^0_{i/\mathrm{EE},t}=
    \pos^0_{i,t} - \pos^0_{\mathrm{EE},t}$ and $\vel^0_{i/\mathrm{EE},t} = \vel^0_{i,t} - \vel^0_{\mathrm{EE},t}$ respectively. For any ordered subset $\mathcal{J}\subseteq\mathcal{I}_V$, define 
\begin{equation} 
\mathbf{h}_t(\mathcal{J}) = \left[ \left(\pos^0_{i/\mathrm{EE},t}\right)_{i\in\mathcal{J}},\, \left(\vel^0_{i/\mathrm{EE},t}\right)_{i\in\mathcal{J}},\, \pos^0_{k_{\mathrm{tip}},t} \right] \in\R^{4|\mathcal{J}|+2} \label{eq:dlo_observation}
\end{equation}
The actor and critic DLO observations are then $\obsdloact_t=\mathbf{h}_t(\mathcal{I}_{\pi})$ and $\obsdlocri_t=\mathbf{h}_t(\mathcal{I}_V)$, respectively. Thus, the actor receives the relative positions and velocities of the
vision-derived DLO keypoints together with the absolute DLO-tip position $\pos^0_{k_{\mathrm{tip}},t}$ in the robot-base frame. At the beginning of each episode, an ordered tuple of distinct task indices
$\mathcal{K}=(k_1,\ldots,k_M)$, with $k_m\in\mathcal{I}_{\pi}$, is selected.
For each $k_m$, a target $\go^0_{G,k_m}\in\R^2$ is sampled from the task
distribution $\mathcal{D}$. Its displacement and distance are
$\mathbf{e}_{k_m,t}=\go^0_{G,k_m}-\pos^0_{k_m,t}$ and
$d_{k_m,t}=\|\mathbf{e}_{k_m,t}\|_2$, respectively. The task observation is
$\obstas_t=[(\mathbf{e}_{k_m,t})_{m=1}^{M}]\!\in\R^{2M}$. This construction provides a common formulation for tracking an arbitrary
fixed number $M$ of DLO keypoints. The task-observation dimension scales with
$M$; therefore, a conventional fixed-dimensional policy is instantiated for
a particular value of $M$. The privileged critic uses the dense DLO observation
$\obsdlocri_t=\mathbf{h}_t(\mathcal{I}_V)$ and receives
\begin{equation}
    \obs^V_t =
    \left[
        \obsrob_t,\,
        \obsdlocri_t,\,
        \obstas_t,\,
        \mathbf{z}_t,\,
        \act_{t-1}
    \right]
    \in\R^{18+4N_V+2M},
    \label{eq:critic_observation}
\end{equation}
where $\mathbf{z}_t\in\R^5$ contains the normalized command-clipping
magnitude and indicators associated with violations of physical boundaries,
including the robot workspace and admissible work area. This asymmetric actor--critic formulation allows the
critic to exploit the full simulated DLO state during training while keeping
the actor observation compatible with real-world deployment. The actor outputs the normalized three-dimensional continuous action $\act_t
    =
    \left[
        {\mathbf{u}^{xy}_t}^{\top},\,
        u^\psi_t
    \right]^\top
    \in[-1,1]^3$, where $\mathbf{u}^{xy}_t=[u^x_t,u^y_t]^\top$
commands the planar EE translation and $u^\psi_t$ commands the in-plane yaw increment. The corresponding Cartesian commands are $\Delta\pos^{0}_{s,t}
    =
    s_{xy}\mathbf{u}^{xy}_t$ and $\Delta\psi_t
    =
    s_{\psi}u^\psi_t$, where $s_{xy}$ and $s_{\psi}$ are fixed translation and rotation scaling
factors, respectively. 

While the preceding formulation supports any fixed number of task-relevant keypoints, the reward considered in the remainder of this section is instantiated for two-point manipulation, i.e., $\mathcal{K}=(k_{\mathrm{tip}},k_{\mathrm{att}})$ and $M=2$. The primary objective is to guide the free DLO tip to its target, after which the EE-attached DLO keypoint is guided to its corresponding target. The reward accounts for goal tracking, $r^{\mathrm{goal}}_t$ $\in\R$, progress toward the currently active objective, $r^{\mathrm{prog}}_t$ $\in\R$, smooth and physically feasible motion, $r^{\mathrm{reg}}_t$ $\in\R$, and terminal or target-exit events, $r^{\mathrm{event}}_t$ $\in\R$:\begin{equation}
    r_t
    =
    r^{\mathrm{goal}}_t
    +
    r^{\mathrm{prog}}_t
    +
    r^{\mathrm{reg}}_t
    +
    r^{\mathrm{event}}_t,
    \qquad
    r_t\in\R.
    \label{eq:reward_decomposition}
\end{equation}
The activation variable for the secondary, EE-attached keypoint objective is defined as $\phi_t
    =
    \mathbb{I}
    \left[
        d_{k_{\mathrm{tip}},t}
        \leq
        \epsilon
    \right]$, where $\epsilon=0.03\,\mathrm{m}$. The goal-tracking reward is $r^{\mathrm{goal}}_t=
-\lambda_{\mathrm{tip}}d_{k_{\mathrm{tip}},t}
-\phi_t\lambda_{\mathrm{att}}d_{k_{\mathrm{att}},t}$,
with $\lambda_{\mathrm{tip}}=\lambda_{\mathrm{att}}=1$. Thus, the tip error is penalized throughout the episode, whereas the error of
the EE-attached keypoint is penalized only after the tip has entered its target region. Defining the positive reduction in the error of keypoint $j$ as $\Delta d^{+}_{j,t}=[d_{j,t-1}-d_{j,t}]_{+}$, positive progress toward the currently active objective is rewarded by
$r^{\mathrm{prog}}_t=\lambda_{\mathrm{prog}}\bigl((1-\phi_t)
\Delta d^{+}_{k_{\mathrm{tip}},t}+\phi_t
\Delta d^{+}_{k_{\mathrm{att}},t}\bigr)$, where
$[x]_{+}=\max(x,0)$ and $\lambda_{\mathrm{prog}}=10$. If the tip
subsequently leaves its target region, $\phi_t$ returns to zero and the reward again prioritizes recovery of the tip. Smooth and physically feasible motion is encouraged by defining
$\Delta\mathbf{u}^{xy}_t=\mathbf{u}^{xy}_t-\mathbf{u}^{xy}_{t-1}$ and
$\Delta u^\psi_t=u^\psi_t-u^\psi_{t-1}$. The corresponding regularization
reward is
$r^{\mathrm{reg}}_t=
-\lambda_{xy}\lVert\Delta\mathbf{u}^{xy}_t\rVert_2
\allowbreak-\lambda_{\psi}|\Delta u^\psi_t|
\allowbreak-\lambda_{\dot{\psi}}\dot{\psi}_t^2
\allowbreak-\lambda_{\mathrm{EE}}\lVert\vel^0_{\mathrm{EE},t}\rVert_2^2
\allowbreak-\lambda_{\mathrm{rel}}
\lVert\vel^0_{k_{\mathrm{att}},t}
-\vel^0_{k_{\mathrm{tip}},t}\rVert_2^2$. The first two terms penalize abrupt changes in the commanded action, the next
two penalize excessive EE yaw rate and translational speed, and the final term
penalizes large relative motion between the two controlled DLO keypoints. In
particular, the EE-velocity and relative-velocity penalties discourage jerky
EE motion and aggressive pulling of the DLO. The corresponding weights are
$\lambda_{xy}=0.05$, $\lambda_{\psi}=0.001$,
$\lambda_{\dot{\psi}}=0.005$, $\lambda_{\mathrm{EE}}=0.25$, and
$\lambda_{\mathrm{rel}}=1$.

The two-point task is considered successful when both controlled keypoints lie
within their respective target regions, i.e.,
$s_t=\mathbb{I}\!\left[
d_{k_{\mathrm{tip}},t}\leq\epsilon
\allowbreak\land
d_{k_{\mathrm{att}},t}\leq\epsilon
\right]$. Let $\tau_t$ denote episode termination and $\ell_t$ indicate that
a controlled keypoint leaves its target region after entering it but before
completion of the combined objective. The event reward is
$r^{\mathrm{event}}_t=
\beta_s s_t
\allowbreak-\beta_T
\mathbb{I}\!\left[\tau_t\land\neg s_t\right]
\allowbreak-\beta_{\mathrm{leave}}\ell_t$,
where $\beta_s=\beta_T=150$ and $\beta_{\mathrm{leave}}=16$. Invalid
termination may result, for example, from a robot-workspace violation or the
DLO leaving the support surface. For single-point tracking,
$\mathcal{K}=(k_{\mathrm{tip}})$; the terms involving
$k_{\mathrm{att}}$ are omitted, such that
$r^{\mathrm{goal}}_t=-\lambda_{\mathrm{tip}}d_{k_{\mathrm{tip}},t}$,
$r^{\mathrm{prog}}_t=\lambda_{\mathrm{prog}}
\Delta d^+_{k_{\mathrm{tip}},t}$, and
$s_t=\mathbb{I}[d_{k_{\mathrm{tip}},t}\leq\epsilon]$.

\subsection{Training Strategy}
\label{subsec:training_strategy}
Policies are trained using PPO~\cite{schulmanppo}, as implemented in MuJoCo Playground~\cite{zakka2025mujocoplayground}. The reward formulation above defines the raw environment reward; within the MP PPO pipeline, observations and scalar rewards are normalized before being used for policy and value-function optimization. The actor and critic are multilayer perceptrons with SiLU activations, comprising two and three hidden layers of $256$ units,
respectively. The policy operates at $10\,\mathrm{Hz}$, while the physics simulation is
integrated at $1\,\mathrm{kHz}$. Each training episode lasts at most 200
control steps (20 s). Framework-specific PPO hyperparameters are selected as described in Sec.~\ref{subsec:hpo}.

For robustness-oriented training, the environment parameters are randomized at the beginning of each episode. The planar sliding-friction coefficient between the DLO and the support surface is uniformly sampled as $\mu_{\mathrm{p}} \sim \mathcal{U}(0.5,0.8)$. Independently, the scale factors for the robot joint stiffness, actuator damping, and passive damping are each uniformly sampled from $\mathcal{U}(0.95,1.05)$. The initial DLO
configuration and the initial EE planar pose are also perturbed at each reset. Disturbance-recovery training additionally
applies an external force at a randomly selected time during the episode. Bending and torsional
stiffness remain fixed because the employed MP plugin does not support their randomization during runtime. At each control step, independent zero-mean Gaussian
noise perturbs actor-observable EE/DLO positions and velocities, yaw, and yaw
rate. Relative quantities are recomputed after perturbation, and a random delay
of at most one control step is applied. The critic receives the corresponding
clean privileged state.

\section{Task Description}
The system comprises a robot and a DLO supported on a planar
work surface, as illustrated in Fig.~\ref{fig:task_description}. The work surface affects the
DLO through contact and friction but is not included in the policy observation.
Furthermore, the task is restricted to slow planar motion, such that dynamic effects are neglected, and the manipulation is modeled as quasi-static at the control level.
One end of the DLO is rigidly grasped
by the robot's EE, and the Cartesian controller restricts the
commanded EE motion to translation in the $xy$-plane and rotation about the
$z$-axis.  In this work, the stacks are trained and evaluated on the following two tasks:
\begin{enumerate*}[label=(\roman*)]
    \item single-point tracking, and 
    \item two-point tracking.
\end{enumerate*}
The single-point task tracks the last DLO keypoint $\pos^0_{k_{\mathrm{tip}},t}$ toward its target $\go^0_{G,k_{\mathrm{tip}}}$, whereas the two-point task additionally tracks the EE-attached DLO keypoint $\pos^0_{k_{\mathrm{att}},t}$ toward its corresponding target $\go^0_{G,k_{\mathrm{att}}}$. For the two-point task, $\go^0_{G,k_{\mathrm{att}}}$ is sampled conditionally on $\go^0_{G,k_{\mathrm{tip}}}$, subject to feasibility of the EE workspace and DLO configuration. Consequently, $\go^0_{G,k_{\mathrm{att}}}$ may lie outside the target sampling region used for the last DLO keypoint. The actor observes $N_\pi=5$ DLO keypoints in both simulation and real-world deployment, while the critic uses a denser privileged representation with $N_V=12$ DLO keypoints during training.
\section{Experimental Evaluation}
\label{sec:experimental_evaluations}
This section examines how computational and representational design choices affect the efficiency, reliability, and transferability of model-free RL for DLO manipulation. 
We first investigate whether accelerating policy optimization or physics simulation provides the greater practical benefit, and then ablate the observation design. 
Using the selected configuration, we then evaluate single- and two-point manipulation under observation perturbations, randomized dynamics, external disturbances, and DLO geometry variation, followed by zero-shot deployment on the physical robot.

\subsection{Computational Acceleration}
To determine where computational acceleration matters most, we compare three complete PPO--simulation stacks. 
The conventional baseline, denoted SB3+mj, combines the PyTorch-based Stable-Baselines3 implementation with CPU-executed MuJoCo~\cite{stable-baselines3,todorov2012mujoco} using four parallel environments. 
SBX+mj replaces the learning backend with the JAX-based Stable-Baselines JAX implementation while retaining the same CPU-based simulation setup, thereby assessing the benefit of accelerating policy optimization alone. 
The third stack uses MuJoCo Playground (MP)~\cite{zakka2025mujocoplayground}, which integrates JAX-based PPO with MJX, a JAX-native MuJoCo simulation backend. This enables both policy optimization and experience generation to be parallelized on the GPU across $1024$ environments.

Because these stacks differ in their learning framework, simulation backend, degree of parallelism, and hardware execution, we evaluate them as complete training systems. 
The comparison therefore investigates whether an accelerated learning backend or massively parallel physics simulation provides the greater practical benefit for data-intensive DLO manipulation.

\subsubsection{Hyperparameter Optimization}
\label{subsec:hpo}
To improve comparability across implementations, a separate hyperparameter sweep was conducted for each configuration using Optuna~\cite{optuna} with a Tree-structured Parzen Estimator. Each trial was evaluated over three tuning seeds using the success-oriented objective $0.90\,\bar{s}_{\mathrm{final}} + 0.10\,\mathrm{AUC} - 0.35\,\sigma(s_{\mathrm{final}})$, where $\bar{s}_{\mathrm{final}}$ and $\sigma(s_{\mathrm{final}})$ are the mean and standard deviation of the final success rate (SR) across seeds, and AUC is the area under the evaluation SR curve. The objective therefore balances final success, learning performance, and robustness across seeds. All sweeps were conducted using eight CPU cores and one NVIDIA A100-SXM4 GPU with 80\, GB of memory. The search spaces and selected incumbent configurations for each training stack are reported in Table~\ref{tab:hyperparam_combined}.

\subsubsection{Performance Comparison} We compare each training stack with its incumbent parameter set by training them from scratch on 5 held-out seeds. The reported metrics comprise the final success rate (Final SR), the area under the SR curve (SR AUC), the final episodic reward, and the area under the reward curve (Reward AUC) across all seeds. 
For the rest of the paper, all runs use 8 CPU cores and one NVIDIA Tesla V100-SXM2 GPU with 32\, GB of memory.
\begin{table}[t]
\caption{PPO search spaces and selected incumbents.}
\centering
\label{tab:hyperparam_combined}
\scriptsize
\setlength{\tabcolsep}{2.2pt}
\renewcommand{\arraystretch}{1.2}
\begin{tabularx}{\columnwidth}{@{}l|X|c|c|c@{}}
\hline
\hline
\textbf{Parameter} & \textbf{Search space} & \textbf{SB3} & \textbf{SBX} & \textbf{MP} \\
\noalign{\hrule height 1pt}
Learning rate & $\{1,2,3,4,5,8,10\}{\times}10^{-4}$ & $8{\times}10^{-4}$ & $8{\times}10^{-4}$ & $8{\times}10^{-4}$ \\
\hline
Update epochs & $10,15,20$; MP: $4,8,12$ & 10 & 10 & 12 \\
\hline
Discount $\gamma$ & $0.985,0.990,0.995$ & 0.990 & 0.985 & 0.990 \\
\hline
GAE lambda & $0.90,0.95,0.97,0.98$ & 0.97 & 0.95 & 0.97 \\
\hline
Clip range & $0.10$--$0.30$ in $0.05$ steps & 0.10 & 0.20 & 0.20 \\
\hline
Entropy coeff. & $\{1,3,10,30\}{\times}10^{-3}$ & 0.010 & 0.030 & 0.020 \\
\hline
Value coeff. & $0.5,0.8,1.0$; MP fixed at $0.5$ & 0.8 & 0.5 & 0.5 \\
\hline
\hline
\end{tabularx}
\end{table}

\subsubsection{Results}
\label{subsec:incumbent}

\begin{figure}[t]
    \centering
    \includegraphics[width=0.95\columnwidth,
    keepaspectratio]{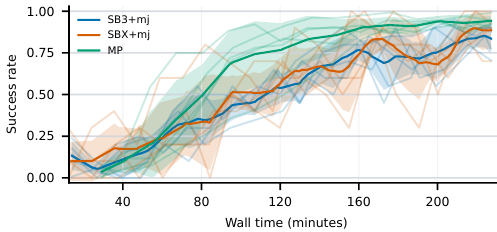}
    \caption{Mean success rate (SR) learning curves of the selected PPO incumbents as a
    function of wall-clock time over seeds 3--7 with shaded bands indicate cross-seed variability.}
    \label{fig:wall_time_training}
\end{figure}

\begin{table}[t]
\scriptsize
\setlength{\tabcolsep}{4pt}
\renewcommand{\arraystretch}{1.2}
    \centering
    \caption{Mean $\pm$ standard deviation of the selected PPO incumbents over five random seeds.}
    \label{tab:incumbent_seed_variability}
    \begin{tabular}{c|c|c|c|c}
        \hline
        \hline
        \textbf{Method} & \textbf{Final SR} & \textbf{SR AUC} & \textbf{Final reward} & \textbf{Reward AUC} \\
        \noalign{\hrule height 1pt}
        SB3+mj & $0.860 \pm 0.108$ & $0.529 \pm 0.048$ & $114.112 \pm 21.857$ & $48.958 \pm 9.904$ \\
        \hline
        SBX+mj & $0.880 \pm 0.110$ & $0.532 \pm 0.069$ & $116.347 \pm 30.666$ & $42.836 \pm 16.035$ \\
    \hline
        MP  & $0.950 \pm 0.032$ & $0.735 \pm 0.077$ & $122.628 \pm 8.116$ & $74.702 \pm 16.157$ \\
    \hline 
    \hline 
    \end{tabular}
  \vspace{-5mm}
\end{table}
All three stacks learn the
task, but exhibit markedly different convergence speeds and cross-seed
stability, as summarized in Table~\ref{tab:incumbent_seed_variability}.
The wall-clock comparison in Fig.~\ref{fig:wall_time_training} highlights the
practical cost of CPU-based simulation. Within the common training budget of
approximately $240$\, min, SB3+mj and SBX+mj attain only moderate success,
whereas MP reaches a mean SR of $0.95$. This difference is consistent
with the substantially greater amount of experience collected by MP through
massive parallelization, using $1024$ MJX environments compared with four
CPU-executed MuJoCo environments. Over the same wall-clock interval, MP
therefore collects approximately $13\times$ more environment interactions.
In summary, the results
confirm that obtaining high success with CPU-based MuJoCo requires a much
larger wall-clock budget than with MP.
Beyond training speed, the seed variability provides further insight into the
reliability of the three configurations. 
Considering the final episode, SB3+mj and SBX+mj exhibit comparable
variability in success, although SBX+mj shows greater dispersion in
reward and SR AUC. MP achieves the highest success and reward
with the lowest variability, while its AUC variability
remains comparable to that of the CPU-based configurations. Taken together,
these results show that CPU-based simulation is considerably less practical for
DLO manipulation training at this scale, while massive parallelization improves
both training efficiency and final policy reliability.
Based on this comparison, MP was selected as the training
stack for the remainder of this work. 
\subsection{Observation-Design Ablation}
\label{subsec:ablation_protocol}
The observation-design ablation was conducted using the single-point tracking task and the selected MP incumbent. Two designs dimensions are examined: \begin{enumerate*}
    [label=(\roman*)]
    \item representing the DLO keypoint states relative to the EE or in the robot base frame and
    \item  using
symmetric or asymmetric actor--critic observations.
\end{enumerate*}
Each configuration is trained from scratch for approximately $20$ million environment steps using two random seeds. 
The PPO hyperparameters, reward formulation, network architecture, and environment settings are held
fixed across all four configurations. Performance is compared using the mean SR curves obtained throughout training. 
\subsubsection*{Results} 
\label{subsec:observation_ablation} 
\begin{figure}[t]
    \centering
    \includegraphics[width=0.95\linewidth]{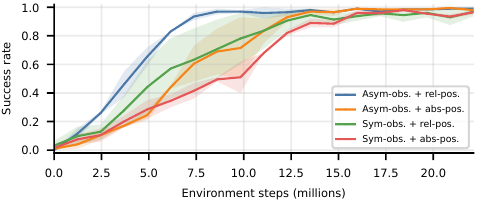}
    \caption{Observation-design ablation using MP. Curves show the mean evaluation success rate (SR), and shaded regions indicate seed variability.}
    \label{fig:ablation}
    \vspace{-5 mm}
\end{figure}

Fig.~\ref{fig:ablation} presents the evaluation SR for the four observation configurations. 
Representing the DLO keypoint states relative to the EE improves learning compared with the corresponding robot-base-frame representation. The asymmetric actor--critic formulation, in which the critic receives privileged simulation information while the actor retains the deployable observation, also performs better than the symmetric formulation. The relative, asymmetric configuration achieves the strongest overall performance in this ablation and is therefore adopted for all subsequent simulation evaluations and real-world experiments.
\subsection{Robustness and Generalization in Simulation}
\label{subsec:sim_eval_protocol}
The system is further evaluated to assess its robustness and generalization aspects. 
In the default configuration, the DLO has a length of $0.30$\,m, and the relative, asymmetric observation
design in Sec.~\ref{subsub:tdef} is used. The reward and termination rules
follow Sec.~\ref{subsub:tdef}, while the policy architecture, control
frequency, and episode horizon follow Sec.~\ref{subsec:training_strategy}.
\subsubsection{Tasks and Evaluation Conditions}
The single-point and two-point task formulations defined in
Sec.~\ref{subsub:tdef} are evaluated. For each task, two principal training
and evaluation conditions are considered.
The \emph{Basic} condition uses nominal simulation parameters and clean actor
observations, without observation noise or delay. The \NDR~condition includes the observation perturbations and delay, and episode-level domain
randomization described in Sec.~\ref{subsec:training_strategy}. For
single-point tracking, a third condition,
\NDRD, additionally evaluates the disturbance-recovery
policy under an external force perturbation applied to $\pos^0_{k_{\mathrm{tip}},t}$ during the
episode. When task conditions are compared directly, the same evaluation seeds,
sampled targets, and randomized initial DLO configurations are reused. The
analysis addresses three questions: 
\begin{enumerate*}
    [label=(\roman*)]
    \item how increasing the objective from one to two tracked points affects performance,
    \item whether single-point tracking remains robust to external perturbations and DLO-length variation,
    \item where failures occur in the more demanding two-point task.
\end{enumerate*}
\subsubsection{Evaluation Repetitions and Metrics}
Each task--condition pair is evaluated using five random evaluation seeds and
$50$ episodes per seed, yielding $250$ episodes in total. These seeds repeatedly
evaluate the same trained policy and therefore capture variability across
targets, initial states, dynamics parameters, and observation perturbations.

The primary metric is the episode success rate (SR) under the task-specific
criterion defined in Sec.~\ref{subsub:tdef}. The reported results include five
seed-specific estimates, the pooled SR over all
$250$ episodes, and its $95\%$ Wilson confidence interval. Unsuccessful
episodes are categorized as timeouts or safety terminations. A timeout is an
unsuccessful episode that reaches the evaluation horizon, whereas a safety
termination results from a table- or workspace-related termination; the total
number of failures is the sum of these two outcomes. For two-point tracking,
failures are further decomposed into three task-level categories:
Type-1 indicates that $\go^0_{G,k_{\mathrm{tip}}}$ was never satisfied;
Type-2 indicates that $\go^0_{G,k_{\mathrm{tip}}}$ was satisfied but the
target $\go^0_{G,k_{\mathrm{att}}}$ was not; and Type-3 indicates that
$\go^0_{G,k_{\mathrm{tip}}}$ was satisfied during the episode but was no longer satisfied
at termination. For single-point tracking, only Type-1 is
applicable.

For the workspace diagnostics, a two-point target pair is classified as
peripheral when either target lies within $0.05$\,m of its 
admissible workspace boundary. Two-point success is also stratified by the
target separation using the intervals below $0.10$\,m,
$[0.10,0.20)$\,m, and at least $0.20$\,m.

In addition to the fixed-length experiments, a separate
single-point policy is evaluated after being trained jointly on DLO lengths of $0.300\,\mathrm{m}$, $0.375\,\mathrm{m}$, and $0.450\,\mathrm{m}$ in simulation. Since all three lengths are included during
training, this experiment measures in-distribution robustness to DLO-length
variation rather than extrapolation to an unseen length. The policy is
evaluated separately at each length using the same five-seed,
$50$-episode-per-seed protocol.
\subsubsection{Results}
\label{subsec:simulation_results}

\begin{figure}[!t]
    \centering
    \includegraphics[width=0.9\columnwidth]{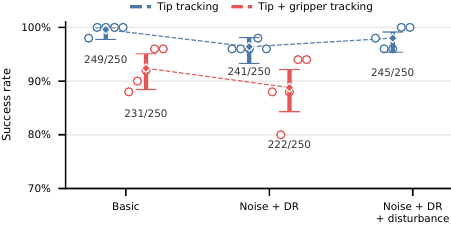}
    \caption{Simulation success rates for single- and two-point tracking. Open circles show estimates
    from each of five evaluation seeds; diamonds and
    error bars show pooled estimates and $95\%$ Wilson intervals. \NDRD~ was evaluated only for
    single-point tracking.}
    \label{fig:wilson_plot}
\vspace{-5mm}
\end{figure}

The results are given in five categories below.
\paragraph{Overall tracking performance}
Table~\ref{tab:simulation_diagnostics} summarizes the episode outcomes and
failure types, and
Fig.~\ref{fig:wilson_plot} reports the corresponding pooled SR and
$95\%$ Wilson intervals. Single-point tracking achieved $99.6\%$
success under \emph{Basic}, $96.4\%$ under \NDR, and
$98.0\%$ when external disturbances were added to \NDR.
Two-point tracking was more demanding, reaching $92.4\%$
success under \emph{Basic} and $88.8\%$ under \NDR. The
remainder of this section analyses the robustness and failure mechanisms behind
these aggregate results.
\begin{table}[t]
    \caption{Simulation outcomes and failure decomposition over 250 episodes per condition.}
    \label{tab:simulation_diagnostics}
    \centering
    {\scriptsize
    \setlength{\tabcolsep}{4pt}
    \renewcommand{\arraystretch}{1.2}
    \resizebox{\columnwidth}{!}{
    \begin{tabular}{l|l|ccc|cccc @{}}
        \hline
        \hline
        & & \multicolumn{3}{c|}{\textbf{Episode outcome}} &
        \multicolumn{4}{c}{\textbf{Failure decomposition}} \\
        \cline{3-9}
        \textbf{Task} & \textbf{Condition} & \textbf{Success} &
        \textbf{Timeout} & \textbf{Safety} & \textbf{Failures} &
        \textbf{Type-1} & \textbf{Type-2} & \textbf{Type-3} \\
        \noalign{\hrule height 1.2pt}
        \multirow{3}{*}{Single-point}
            & \emph{Basic}                    & 249 & 1  & 0 & 1  & 1  & -- & 0 \\
        \cline{2-9}
            & \NDR~               & 241 & 5  & 4 & 9  & 9  & -- & 0 \\
        \cline{2-9}
            & \NDRD~ & 245 & 3  & 2 & 5  & 5  & -- & 0 \\
        \hline
        \multirow{2}{*}{Two-point}
            & \emph{Basic}                    & 231 & 16 & 3 & 19 & 9  & 1 & 9 \\
        \cline{2-9}
            & \NDR~              & 222 & 28 & 0 & 28 & 19 & 2 & 7 \\
        \hline
        \hline
    \end{tabular}}
    }
    \vspace{-2mm}
\end{table}
\paragraph{Single-Point Tracking Robustness}
For single-point tracking, \NDR~ reduced the SR by $3.2$
percentage points relative to \emph{Basic}, while performance remained high when an
external disturbance was additionally applied. During training, it was observed
that the applied force occasionally displaced $\pos^0_{k_{\mathrm{tip}},t}$ toward the target,
helping the policy reach configurations that would otherwise have been more
difficult to attain. This incidental effect may partly explain the slightly
higher SR under disturbance. Since the \NDR~and \NDRD~evaluations use identical targets
and randomized initial states, their outcomes can be compared episode by
episode. Performance was unchanged in $238$ of the $250$ episodes, while eight episodes
changed from failure to success and four changed from success to failure under
disturbance. Thus, the slightly higher SR under the disturbance condition should
be interpreted as evidence of retained robustness, rather than evidence that the
perturbation systematically improves performance.

\paragraph{Two-Point Tracking Robustness}
Simultaneously satisfying the targets $\go^0_{G,k_{\mathrm{tip}}}$ and $\go^0_{G,k_{\mathrm{att}}}$  was more demanding than
single-point tracking. Under \NDR, the two-point SR decreased by
$3.6$ percentage points relative to \emph{Basic}. Because both conditions use the
same $250$ target pairs, their outcomes can also be compared episode by
episode: $213$ pairs succeeded under both conditions, $18$ succeeded only
under \emph{Basic}, $9$ succeeded only under \NDR, and $10$ failed under both.
Thus, \NDR~produces a modest net degradation rather than a uniform loss of
capability across the sampled task configurations.
\paragraph{Failure and Workspace Diagnostics}
Table~\ref{tab:simulation_diagnostics} complements the aggregate SR
with a decomposition of the failure modes for both tracking tasks. In the
single-point task, an episode terminated once $\pos^0_{k_{\mathrm{tip}},t}$ satisfied the
required threshold. Consequently, all failures are Type-1: Type-2 is not
applicable, and Type-3 cannot occur at termination. The \emph{Basic}, \NDR, and
\NDRD~conditions produce $1$, $9$, and $5$ Type-1 failures, respectively.

For two-point tracking, the $19$ failures under \emph{Basic} are distributed as $47.4\%$ Type-1, $5.3\%$ Type-2, and $47.4\%$ Type-3 failures. Under \NDR, Type-1 is the dominant failure mode, accounting for $67.9\%$ of the failures, whereas Type-2 and Type-3 failures account for $7.1\%$ and $25.0\%$, respectively. All failures under \NDR~are caused by timeouts, whereas \emph{Basic} produced $16$ timeouts and
three safety terminations. 

\begin{figure}[t]
    \centering
    \includegraphics[width=0.95\columnwidth]
    {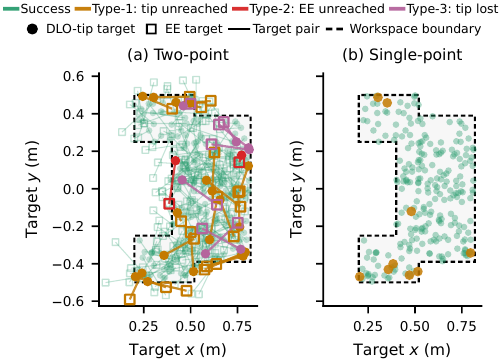}
    \caption{\NDR~outcomes across the two-point (a) and single-point
(b) target workspaces. Colors indicate episode outcomes, markers
identify target types, and black outlines delimit the admissible
workspace.}
\vspace{-5 mm}
\label{fig:workspace_diagnostics}
\end{figure}

Fig.~\ref{fig:workspace_diagnostics} shows the spatial distribution of the
\NDR~outcomes. For single-point tracking, Fig.~\ref{fig:workspace_diagnostics}(b)
maps the $241$ successful episodes and $9$ Type-1 failures over the sampled
DLO's target workspace. For two-point tracking,
Fig.~\ref{fig:workspace_diagnostics}(a) shows that the failures are not distributed
uniformly over the target-pair geometry.

The peripheral target pair is defined as one for which either target lies within $0.05\,\mathrm{m}$ of its corresponding admissible workspace boundary. Under
\NDR, $78.6\%$ of failures occur for peripheral pairs, although such
pairs constitute only $44.4\%$ of the $250$ episodes. The corresponding failure
rate near the workspace boundaries is $19.8\%$, compared with
$4.3\%$ in the interior. The same tendency is weaker under \emph{Basic},
with failure rates of $10.8\%$ and $5.0\%$, respectively.

Relative target geometry also shows a descriptive association with difficulty. Under \NDR, the observed SR was \(92.7\%\) for target separations below \(0.10\,\mathrm{m}\), \(90.9\%\) for separations in \([0.10,0.20)\,\mathrm{m}\), and \(83.1\%\) for separations of at least \(0.20\,\mathrm{m}\). Among the 28 failures, the median target separation was
$0.186\,\mathrm{m}$ for Type-1 ($n=19$), $0.136\,\mathrm{m}$ for Type-2
($n=2$), and $0.203\,\mathrm{m}$ for Type-3 ($n=7$).
The results show that the loss of robustness
is concentrated near the workspace limits and in geometrically demanding target-pair configurations.
\paragraph{Variable-Length Policy Performance}
A separate single-point policy trained jointly on DLO lengths of
$0.300\,\mathrm{m}$, $0.375\,\mathrm{m}$\, and $0.450\,\mathrm{m}$ was evaluated.

\begin{figure}[t]
    \centering
    \includegraphics[width=0.9\columnwidth]
    {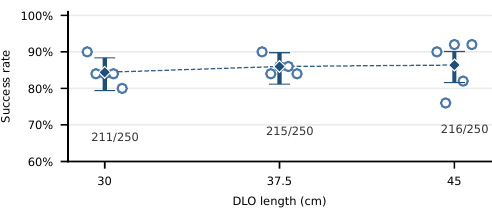}
    \caption{Single-point success across the three trained DLO
lengths.}
    \label{fig:generalised_dlo}
    \vspace{-4mm}
\end{figure}

As shown in Fig.~\ref{fig:generalised_dlo}, the policy successfully completed the task in $84.4\%$ of the trials for a DLO length of $0.300$\,m. For DLO lengths of $0.375\,\mathrm{m}$ and $0.450\,\mathrm{m}$, the corresponding SR are $86.0\%$, and $86.4\%$, respectively. The pooled SRs span only $2.0$
percentage points, and their Wilson confidence intervals overlap, indicating comparable performance across the three trained lengths.
\subsection{Zero-Shot Real-World Evaluation}
\label{subsec:real_world_protocol}
The real-world platform consists of a Franka Emika Panda robot mounted beside
a fixed table. One end of the DLO remains grasped by the EE throughout the
experiment, while the opposite end is free. An Intel RealSense D435i camera is
mounted above the workspace and provides images to the perception pipeline
described in Sec.~\ref{subsec:vision} at $30\,\mathrm{Hz}$. The camera frame is
registered with the robot-base frame using ArUco-based extrinsic calibration,
and the complete perception and control pipeline is integrated through the
Robot Operating System (ROS). The YOLO26-seg model described in Sec.~\ref{subsec:vision} returns $N_\pi=5$ keypoints.
The policies are deployed zero-shot, without real-world fine-tuning or policy network changes. The actor operates at the same
$10$\, Hz control frequency used in simulation.

\subsubsection{Primary Evaluation}
The primary real-world evaluation uses a $0.004$\,m diameter silicone catheter and
comprises $10$ trials for each of the three evaluated tasks: single-point
tracking, disturbance recovery, and two-point tracking. All trials conducted
as part of this primary evaluation are included in the reported results.

At the beginning of each trial, the robot is returned to its nominal
configuration, the catheter is placed in a new initial configuration, and the
required target position or target pair is sampled within the admissible
real-world workspace. A trial is considered successful when the corresponding
task criterion defined in Sec.~\ref{subsub:tdef} is satisfied within
the episode horizon. For successful trials, the task-completion time is recorded. In two-point tracking, we separately record
the time at which the $\go^0_{G,k_{\mathrm{tip}}}$ criterion is first satisfied and the time at which
the combined objective ($\go^0_{G,k_{\mathrm{att}}}$ and $\go^0_{G,k_{\mathrm{tip}}}$) is completed. Completion times are summarized
using the median and interquartile range over successful trials. Unsuccessful
trials are assigned the Type-1--Type-3 categories defined in
Sec.~\ref{subsec:sim_eval_protocol}, where applicable.
\begin{figure*}[!t]
\centering
\begin{overpic}[
width=\linewidth
]{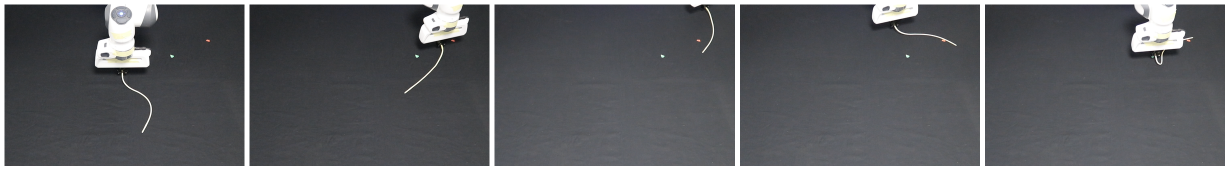}
    \put(0.5,1.2){\snapshotlabel{0 s}}
    \put(20.5,1.2){\snapshotlabel{4 s}}
    \put(40.5,1.2){\snapshotlabel{8 s}}
    \put(60.5,1.2){\snapshotlabel{12 s}}
    \put(80.5,1.2){\snapshotlabel{16 s}}
\end{overpic}
\caption{Snapshots of a successful real-world two-point experiment. The
green and red points denote the targets of the EE-attached and tip
keypoints, $\go^0_{G,k_{\mathrm{att}}}$ and
$\go^0_{G,k_{\mathrm{tip}}}$, respectively.}
\label{fig:snapshots}
\end{figure*}
\subsubsection{Perception Performance}
The fine-tuned YOLO26-seg model achieved a mask precision of $98.81\%$,
recall of $99.13\%$, F1-score of $98.97\%$, mAP@0.5 of $99.48\%$, and
mAP@0.5:0.95 of $79.30\%$, with an inference rate of $29.8$\,FPS.
\begin{table}[]
    \centering
    \caption{Zero-shot real-world results with median completion times.}
    \label{tab:real_world_results}
    \scriptsize
    \setlength{\tabcolsep}{2.2pt}
    \renewcommand{\arraystretch}{1.15}%
    \begin{tabular}{@{}l c c c@{}}
        \hline
        \hline
        \textbf{Task} & \textbf{Success} &
        \textbf{Time [IQR] (s)} & \textbf{Failure type} \\
        \noalign{\hrule height 1.2pt}
        Single-point & 9/10 & 9.30 [6.70, 14.40] &
        Type-1 \\
        \hline
        Disturbance & 10/10 & 6.55 [4.85, 8.32] &
        -- \\
        \hline
       \multirow[l]{2}{*}{Two-point} &
        \multirow[l]{2}{*}{8/10} &
         $\go^0_{G,k_{\mathrm{tip}}}$: 8.55 [7.19, 12.30] &
        \multirow[c]{2}{*}{Type-3} \\
        & &
        Task: 9.05 [7.80, 12.47] &
        \\
        \hline
        \hline
    \end{tabular}%
    \vspace{-5mm}
\end{table}
Table~\ref{tab:real_world_results} summarizes the primary real-world
evaluation. The policies succeeded in $9/10$, $10/10$, and $8/10$ trials for
single-point tracking, disturbance recovery, and two-point tracking, respectively.
\subsubsection{Completion Time and Two-point Coordination}
The conditional median completion times were $9.30$\,s for single-point
tracking, $6.55$\,s for disturbance recovery, and $9.05$\,s for two-point
tracking. In the $8$ successful two-point trials, the target for $\pos^0_{k_{\mathrm{tip}},t}$ was
satisfied after a median of $8.55$\,s [IQR: $7.19$--$12.30$\,s]. The median
delay between satisfaction of the $\pos^0_{k_{\mathrm{tip}},t}$ target and completion of the
two-point objective was $0.56$\,s [IQR: $0.17$--$2.02$\,s]. A time-lapse of
one successful two-point trial is shown in Fig.~\ref{fig:snapshots}. The
sequence shows how the robot first satisfies the target for $\pos^0_{k_{\mathrm{tip}},t}$
at $\go^0_{G,k_{\mathrm{tip}}}$ and then $\pos^0_{k_{\mathrm{att}},t}$ at
$\go^0_{G,k_{\mathrm{att}}}$. In both unsuccessful two-point trials, $\pos^0_{k_{\mathrm{tip}},t}$ reached
but subsequently left its target before the two-point objective was completed. The unsuccessful single-point trial did not reach its target, whereas all disturbance-recovery trials were successful.
\subsubsection{Failure Modes}
The observed real-world failures were mainly associated with perception loss, material effects, and latency. Robot-induced occlusions occasionally degraded the markerless DLO
keypoint estimates; the policy could recover when valid observations returned
near the goal, but longer occlusions led to incorrect actions. The highly
compliant silicone catheter sometimes bent sharply and released stored elastic energy, causing sudden deflections in compact
two-point configurations. Finally, perception delay occasionally made the
policy overshoot the target, although training with observation noise and
delay reduced this effect and allowed recovery in several trials.

\section{Conclusion and Future Work}

This work presented an empirical study of model-free RL for planar DLO manipulation, focusing on computational acceleration, observation design, robustness, and zero-shot sim-to-real transfer. GPU-parallel simulation enabled PPO to collect substantially more experience within the same wall-clock budget than CPU-based training stacks, resulting in faster convergence, higher final success, and lower cross-seed variability. In contrast, accelerating the learning backend alone provided limited practical benefit because the dominant bottleneck remained the DLO physics simulation. The ablation study further showed that relative DLO representations and asymmetric actor--critic observations improve learning efficiency while keeping the deployed policy compatible with vision-based state estimates.

Combined with observation noise and delay, domain randomization, and disturbance-recovery training, the resulting policies transferred to the physical robot without real-world fine-tuning and achieved high success in single-point, disturbed, and two-point manipulation tasks. 
Future work will consider more complex shape-control tasks, three-dimensional manipulation, and improved generalization to unseen DLOs, environments, and task configurations.






\bibliographystyle{ieeetr}
\bibliography{reference}

@article{stable-baselines3,
  author  = {Antonin Raffin and Ashley Hill and Adam Gleave and Anssi Kanervisto and Maximilian Ernestus and Noah Dormann},
  title   = {Stable-Baselines3: Reliable Reinforcement Learning Implementations},
  journal = {Journal of Machine Learning Research},
  year    = {2021},
  volume  = {22},
  number  = {268},
  pages   = {1-8},
  url     = {http://jmlr.org/papers/v22/20-1364.html}
}

@article{todorov2012mujoco,
  title={MuJoCo: A physics engine for model-based control},
  author={Emanuel Todorov and Tom Erez and Yuval Tassa},
  journal={2012 IEEE/RSJ International Conference on Intelligent Robots and Systems},
  year={2012},
  pages={5026-5033},
  url={https://api.semanticscholar.org/CorpusID:5230692}
}

@misc{zakka2025mujocoplayground,
title = {MuJoCo Playground: An open-source framework for GPU-accelerated robot learning and sim-to-real transfer.},
  author = {Zakka, Kevin and Tabanpour, Baruch and Liao, Qiayuan and Haiderbhai, Mustafa and Holt, Samuel and Luo, Jing Yuan and Allshire, Arthur and Frey, Erik and Sreenath, Koushil and Kahrs, Lueder A. and Sferrazza, Carlo and Tassa, Yuval and Abbeel, Pieter},
  year = {2025},
  publisher = {GitHub},
  url = {https://github.com/google-deepmind/mujoco_playground}
}

@InProceedings{nico_rlx,
      title={RL-X: A Deep Reinforcement Learning Library (not only) for RoboCup}, 
      author={Nico Bohlinger and Klaus Dorer},
      year={2023},
      eprint={2310.13396},
      archivePrefix={arXiv},
      primaryClass={cs.RO},
      url={https://arxiv.org/abs/2310.13396}, 
}

@misc{optuna,
      title={Optuna: A Next-generation Hyperparameter Optimization Framework}, 
      author={Takuya Akiba and Shotaro Sano and Toshihiko Yanase and Takeru Ohta and Masanori Koyama},
      year={2019},
      eprint={1907.10902},
      archivePrefix={arXiv},
      primaryClass={cs.LG},
      url={https://arxiv.org/abs/1907.10902}, 
}

@misc{schulmanppo,
      title={Proximal Policy Optimization Algorithms}, 
      author={John Schulman and Filip Wolski and Prafulla Dhariwal and Alec Radford and Oleg Klimov},
      year={2017},
      eprint={1707.06347},
      archivePrefix={arXiv},
      primaryClass={cs.LG},
      url={https://arxiv.org/abs/1707.06347}, 
}

@misc{jocher2026ultralytics,
      title={Ultralytics YOLO26: Unified Real-Time End-to-End Vision Models}, 
      author={Glenn Jocher and Jing Qiu and Mengyu Liu and Shuai Lyu and Fatih Cagatay Akyon and Muhammet Esat Kalfaoglu},
      year={2026},
      eprint={2606.03748},
      archivePrefix={arXiv},
      primaryClass={cs.CV},
      url={https://arxiv.org/abs/2606.03748}, 
}

@misc{long2026self,
      title={Self-supervised Physics-Informed Manipulation of Deformable Linear Objects with Non-negligible Dynamics}, 
      author={Youyuan Long and Gokhan Solak and Sara Zeynalpour and Heng Zhang and Arash Ajoudani},
      year={2026},
      eprint={2602.03623},
      archivePrefix={arXiv},
      primaryClass={cs.RO},
      url={https://arxiv.org/abs/2602.03623}, 
}

@inproceedings{qin2019cable,
  author={Qin, Yili and Escande, Adrien and Yoshida, Eiichi},
  booktitle={2019 IEEE/SICE International Symposium on System Integration (SII)}, 
  title={Cable Installation by a Humanoid Integrating Dual-arm Manipulation and Walking}, 
  year={2019},
  volume={},
  number={},
  pages={98-103},
  doi={10.1109/SII.2019.8700448}}

@ARTICLE{qin2023dual,
  author={Qin, Yili and Escande, Adrien and Kanehiro, Fumio and Yoshida, Eiichi},
  journal={IEEE Robotics and Automation Letters}, 
  title={Dual-Arm Mobile Manipulation Planning of a Long Deformable Object in Industrial Installation}, 
  year={2023},
  volume={8},
  number={5},
  pages={3039-3046},
  doi={10.1109/LRA.2023.3264779}}

@article{yin2021modeling,
author = {Hang Yin  and Anastasia Varava  and Danica Kragic },
title = {Modeling, learning, perception, and control methods for deformable object manipulation},
journal = {Science Robotics},
volume = {6},
number = {54},
pages = {eabd8803},
year = {2021},
doi = {10.1126/scirobotics.abd8803},
URL = {https://www.science.org/doi/abs/10.1126/scirobotics.abd8803},
eprint = {https://www.science.org/doi/pdf/10.1126/scirobotics.abd8803}}

@inproceedings{laezza2021shape,
  author    = {Laezza, Rita and Karayiannidis, Yiannis},
  title     = {Learning Shape Control of Elastoplastic Deformable Linear Objects},
  booktitle = {2021 IEEE International Conference on Robotics and Automation (ICRA)},
  pages     = {4438--4444},
  year      = {2021},
  doi       = {10.1109/ICRA48506.2021.9561984},
  url       = {https://doi.org/10.1109/ICRA48506.2021.9561984}
}

@inproceedings{zakaria2022robotic,
  author={Daniel Zakaria, Mélodie Hani and Aranda, Miguel and Lequièvre, Laurent and Lengagne, Sébastien and Corrales Ramón, Juan Antonio and Mezouar, Youcef},
  booktitle={2022 IEEE 18th International Conference on Automation Science and Engineering (CASE)}, 
  title={Robotic Control of the Deformation of Soft Linear Objects Using Deep Reinforcement Learning}, 
  year={2022},
  volume={},
  number={},
  pages={1516-1522},
  doi={10.1109/CASE49997.2022.9926667}}

@article{daniel2024multiac,
  author={Daniel, Mélodie and Magassouba, Aly and Aranda, Miguel and Lequièvre, Laurent and Corrales Ramón, Juan Antonio and Iglesias Rodriguez, Roberto and Mezouar, Youcef},
  journal={IEEE Robotics and Automation Letters}, 
  title={Multi Actor-Critic DDPG for Robot Action Space Decomposition: A Framework to Control Large 3D Deformation of Soft Linear Objects}, 
  year={2024},
  volume={9},
  number={2},
  pages={1318-1325},
  doi={10.1109/LRA.2023.3342672}}

@article{laezza2024offline,
      title={Offline Goal-Conditioned Reinforcement Learning for Shape Control of Deformable Linear Objects}, 
      author={Rita Laezza and Mohammadreza Shetab-Bushehri and Gabriel Arslan Waltersson and Erol Özgür and Youcef Mezouar and Yiannis Karayiannidis},
      year={2024},
      eprint={2403.10290},
      archivePrefix={arXiv},
      primaryClass={cs.RO},
      url={https://arxiv.org/abs/2403.10290}, 
}

\end{document}